# Volume and leaf area calculation of cabbage with a neural network-based instance segmentation

Nils Lüling[*1], David Reiser[1], H.W. Griepentrog[1]
[1]*University of Hohenheim, Institute of Agricultural Engineering, Garbenstr. 9, D-70599, Stuttgart, Germany*
*nils.lueling@uni-hohenheim.de

**Abstract**

Fruit size and leaf area are important indicators for plant health and are of interest for plant nutrient management, plant protection and harvest. In this research, an image-based method for measuring the fruit volume as well as the leaf area for cabbage is presented. For this purpose, a mask region-based convolutional neural network (Mask R-CNN) was trained to segment the cabbage fruit from the leaves and assign it to the corresponding plant. The results indicated that even with a single camera, the developed method can provide a calculation accuracy of fruit size of 92.6% and an accuracy of leaf area of 89.8% on individual plant level.

**Keywords:** Mask-RCNN, cabbage, instance segmentation, structure from motion

**Introduction**

The importance of sensor based analysis of agricultural crops increases as more and more processes on farms become automated. Plant growth and plant stage are decisive parameters for planning crop protection and fertilization. Furthermore, information about the yield as well as the quality of the crops can be used for harvest logistics and sales. With the increasing development in the areas of computer vision, former time-consuming processes of manual field measurements can become automated.
Instance segmentation of camera images provides the possibility to recognize individual plants and to segment leaves and fruits. From segmented areas in the image, the crop volume and the leaf area (LA) of each individual leaf can be calculated through the additional knowledge of depth information. Due to its large and highly overlapping leaves, cabbage is a difficult plant to analyse. Especially in late growth stages, both the cabbage head and a large part of the cabbage leaves are covered. This situation causes problems in the analysis of individual cabbage plants.
There are already a variety of different methods for calculating LA, using different types of methods and sensors. Attempts with different kinds of depth sensors (Reiser et al, 2019, Vázquez-Arellano et al., 2018) and multispectral sensors (Liu et al, 2016) as well as different types of LA calculations (Campillo et al, 2010, Olfati et al, 2010) have been carried out. For an automated calculation of the leaf area of cabbage at a late growth stage with overlapping leaves, both depth information and an analysis of individual leaves are required. Neural networks provide good possibilities for the automated analysis of plants (Zhang et al, 2020). In order to measure individual leaves and fruit volume, neural network-based instance segmentation is necessary (Xu et al, 2018). For a perfect analysis of the plant, not only the visible leaf area is interesting, but the total leaf area (Zheng &

Moskal, 2009). So far, no research is known to the authors that can automatically calculate the total leaf area and the crop volume of cabbage using camera images.

The aim of this paper is to describe a non-destructive approach to measure total leaf area and fruit volume of individual cabbage plants. For this purpose, the cabbage head, the individual leaves and the entire plant were segmented by a neural network-based instance segmentation using colour (RGB) images. By capturing depth information through a structure from motion (SfM) method, the cabbage volume as well as the total leaf area including the non-visible leaf area were calculated.

**Material and methods**

Dataset recording and ground truth estimation
The dataset of images was recorded near Stuttgart, Germany, two weeks before harvest. A standard GoPro camera (Hero 7, GoPro Inc., San Mateo, CA, USA) was used to capture the training images from a vertical perspective. There was no shading of the images or artificial exposure. To produce high quality depth images, the camera was moved across the rows at a recording frequency of 60 Hz and a resolution of 1920x1440, with a constant height of 90 cm and a constant speed of 1 m/s. The cabbage was a storage cabbage of the cultivar Storidor, that was grown with a row width of 60 cm and a planting distance of 60 cm. As no herbicides were used on the analysed parcel, there was high weed pressure during the recordings that also covered parts of the cabbage (Fig.1).

To validate the calculated cabbage volumes and leaf areas, 10 random picked cabbage plants of the dataset were harvested and measured. To determine the cabbage volume, the circumference of the cabbage head was measured. The volume (V) was estimated as:

$$V = \frac{4}{3} * \pi * Radius^3 \quad (1)$$

This calculated volume fits to the real volume with a correlation coefficient of 0.88 (Radovich & Kleinhenz, 2004) by assuming a spherical shape. To determine the leaf area, all leaves were removed and spread out on a plane and photographed. By segmenting the leaves, the leaf area was recorded in pixels. To convert the pixels into the ground truth leaf area, a reference object of known size was placed on the same plane as the leaves.

Depth image calculation
To create a complementary depth image for each colour image, a SfM method was used (Fig. 1). A process consisting of four steps was necessary to calculate the depth information.
1. The first step was to detect speeded-up-robust-features (SURF) in both images (Bay et al, 2008) and find the matching features (Lowe, 2004).

2. By knowing the position and orientation of the matched features, the images were rectified. The first step of rectification was the calculation of the fundamental matrix. With the fundamental matrix, the transformation matrices for the rectification of the images were calculated (Hartley & Zisserman, 2003).

3. The next step was a disparity measurement using a semi-global matching procedure that generates a disparity image from the two orientated images (Hirschmüller, 2008). Decreasing disparity means that the point is located further away from the camera.

4. The first three steps were carried out both for the following picture and for the previous picture out of the video stream. As a result, two disparity images are available for the initial image. If no disparity could be calculated for a subsequent pixel in the following picture, the disparity from the previous image was used at this pixel coordinate. This step increased the quality of the depth image, especially if the camera was moving at higher speed.

5. By calculating the relative camera positions, the disparity values were standardised. Through the unification, the disparity values were transferred into distance values via a scaling factor, related to sensor size and focal length of the camera.

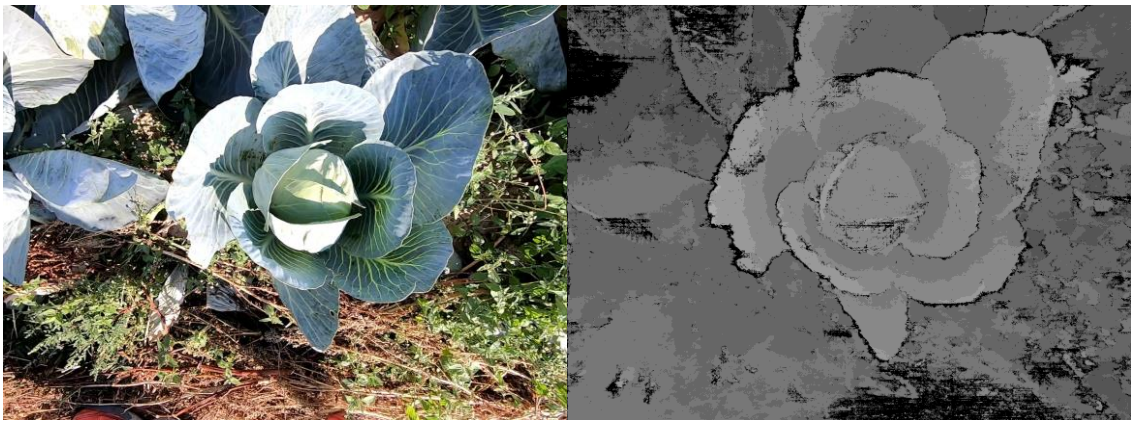

Figure 1: Comparison of colour image and structure from motion depth image.

Instance segmentation

The colour information was used as input for a Mask R-CNN with a Resnet-101 backbone to detect the cabbages and single leaves of the plants. The used Mask R-CNN (Abdulla, 2019) is mainly based on the libraries Tensorflow GPU 1.3.0 (Tensorflow, 2021), Keras 2.0.8 (Keras, 2021) and Python (Python, 2021). The calculation of the leaf area and the cabbage volume, as well as the depth image were implemented with MATLAB (Matlab R2020a, The MathWorks Inc., Natick, MA, USA). The complete training as well as the following steps were carried out with a computer containing a 12-core processor (AMD Ryzen Threadripper 2920X 64 GB RAM, Advanced Micro Devices Inc., Santa Clara, CA, USA) and an 8 GB graphics card (GeForce RTX 2070 Super, Nvidia corporate, Santa Clara, CA, USA).

The training dataset consists of 300 images and the validation dataset consists of 30 images with a resolution of 1024x1024. To avoid overfitting the network, different image augmentation techniques (shift, rotation, scaling, mirroring) were used, which has enlarged the training data set depending on the number of epochs. With 25 epochs, a training data set of 7500 augmented images was created. Furthermore, a transfer learning approach based on the Microsoft common objects in context (COCO) data set was applied to achieve usable results even with a comparatively small data set. The training dataset were trained in three classes (Plant, Crop, Leaf) over 25 epochs, with stochastic gradient descent, an image batch size of one and a learning rate of 0.0001.

Volume and leaf area estimation

For the calculation of the cabbage volume, the cabbage head area segmented by the Mask R-CNN was used. The number of pixels of the segmented area indicates the cross-section of the cabbage head in pixels (Fig.4 a). The diameter in pixels of the cabbage can be calculated via the cross-section of the detected area.

Since the conversion from pixels to cm depends on the distance to the camera, the detected cabbage diameter in pixels was calculated at the height of the cross-section (H) of the cabbage. For this purpose, the radius of the cabbage was subtracted from the height of the centre of the cabbage. With the height of the cabbage centre and a converting constant (k), it was possible to convert the pixel values to centimetres with:

$$r_{cm} = r_{px} * \frac{k}{H_{px}} \qquad (2)$$

By knowing the actual cabbage head radius and using Equation 1, the actual cabbage volume was estimated. By segmenting the entire cabbage plant, the plant-specific leaves were assigned. When more than 50% of a leaf was in the segmentation area of the whole plant, the leaf got assigned to the plant. The problem for calculating the leaf area out of images was that a large part of the leaf area was hidden and had to be estimated. The instance segmentation only provided the visible leaf area. Knowing the approximate leaf attachment below the cabbage fruit in the middle and the outer edges of a leaves, the length of the leaves and therefore the leaf area were determined (Olfati et al, 2010). The calculation was carried out in four steps.

1. In the first step, the three-dimensional coordinate of the leaf attachment (p) was calculated. The two-dimensional centre of the cabbage head was determined, as well as the approximate attachment point of the leaf via the knowledge of the diameter of the cabbage head (Fig.2).
2. In the second step, the two-dimensional coordinate of the segmented leaf (q) which has the greatest two-dimensional distance to the centre was detected. To compensate for undulations of the leaf, the maximum height of the leaf was used as the Z coordinate.

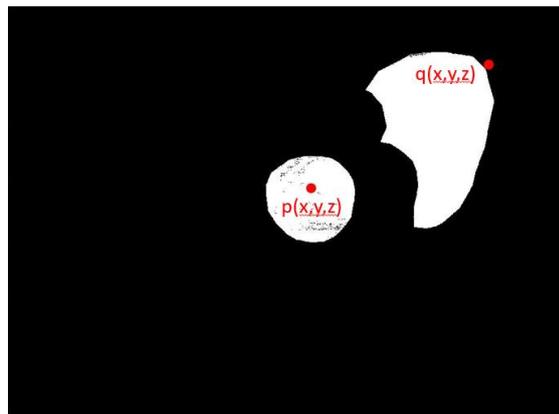

Figure 2: Binary image of the segmented cabbage head and leaf with cabbage centre p and leaf end point q.

3. The three-dimensional euclidian distance (L) between the leaf base (p), which was located at the centre of the cabbage below the cabbage head, and the end of the leaf (q) was then calculated (Fig.3, Eq. 3).

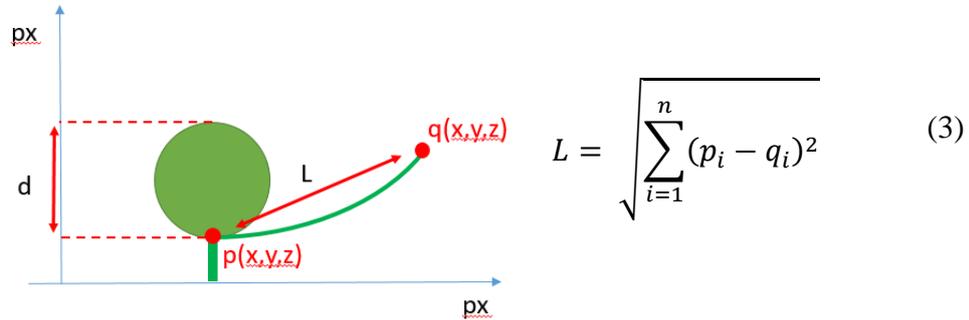

$$L = \sqrt{\sum_{i=1}^{n}(p_i - q_i)^2} \quad (3)$$

Figure 3: Sketch of the side-view of the cabbage head and leaf with measuring points

4. In the last step, the distance (L) was used to estimate the leaf area by a developed equation (Eq. 4). The equation consisted of two parts. In the first part, half of the average length of all leaves was estimated and in the second part, the leaf length (L) was multiplied by a constant parameter.

$$LA = \frac{\frac{1}{n}\sum_{i=1}^{n} d_i}{2} + 8.3 * L \quad (4)$$

**Results and Discussion**

The trained Mask-RCNN performed well for detecting the cabbage plants and cabbage heads. Figure 4 (a) shows the segmentation of the cabbages with a mean average precision (mAP) of 0.81. Figure 4 (b) shows the segmentation of the whole cabbage plant with a general segmentation accuracy of 0.82 mAP and Figure 4 (c) shows the segmentation of all cabbage leaves with an overall accuracy of 0.47 mAP (COCO). The deviations in the segmentation precision of the leaves, are due to large variations of the leaves in shape, colour and size, which makes precise segmentation by a Mask R-CNN difficult.

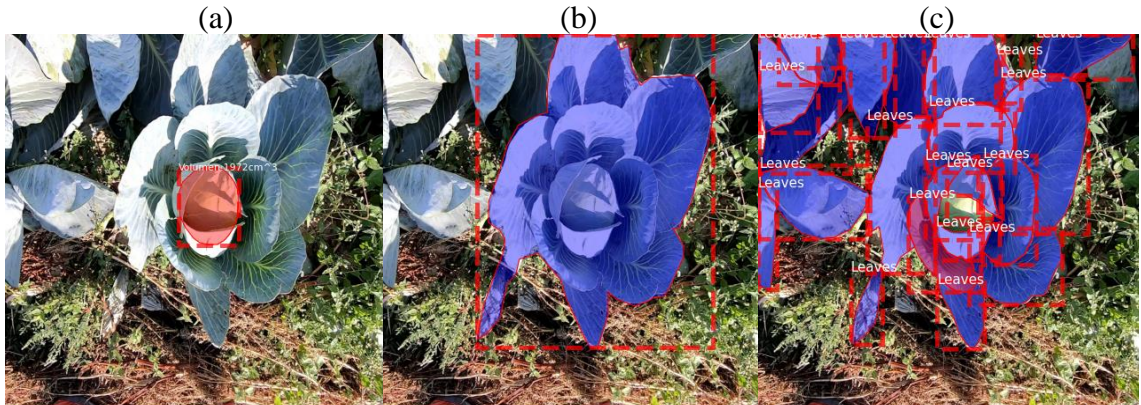

Figure 4: Segmentation of the cabbage head, the cabbage plant and all cabbage leaves.

Cabbage volume

Table 1 shows the measured, calculated and detected results of the cabbage diameter. An average precision of the ground truth diameter of 92.6 % and a precision of the detected diameter of 94.9 % had been achieved. The deviations in the calculated diameter were mainly due to annotation errors of the ground truth data set. Since the real cabbage surface is often obscured by leaves, the cabbage outline had to be estimated when creating the ground truth data.

Table 1. Comparison of measured diameter, calculated ground truth diameter and detected diameter in regard to calculated results.

| Cabbage plant (n) | Measured diameter (cm) | Calc. gTruth diameter image (cm) | Detected diameter (cm) |
| --- | --- | --- | --- |
| 1 | 15.28 | 15.50 | 16.34 |
| 2 | 18.78 | 19.20 | 18.59 |
| 3 | 17.51 | 16.90 | 17.30 |
| 4 | 18.14 | 18.26 | 19.09 |
| 5 | 18.46 | 18.41 | 18.73 |
| 6 | 17.83 | 15.96 | 16.31 |
| 7 | 16.23 | 14.75 | 15.18 |
| 8 | 18.78 | 15.44 | 16.96 |
| 9 | 19.10 | 16.40 | 17.78 |
| 10 | 16.23 | 13.82 | 15.23 |
| Mean precision (%) | | 92.6 | 94.9 |

LA calculation

Table 2 shows the ground truth number of cabbage leaves, as well as the visible and detected ones. For the calculation of the leaf area, only the leaves visible from the camera perspective were evaluated. For the evaluated cabbage plants up to 5 leaves were not visible. This may be due to the heavy overlapping of the leaves as well as the high weed pressure that was present in the field. To calculate the correct leaf area, a correction factor could be included to estimate non-visible leaves. However, this problem is only expected in late growth stages, where a large proportion of the leaves are overlapping. Table 2 shows the comparison of the total leaf area with the visible leaf area and with the leaf area calculated and detected over the leaf length.

With the method created for calculating the leaf area, an average precision of 89.8 % from the real leaf area was achieved. The equation was adapted to Olfati et al, 2010 with the mean value of all leaf lengths and the calculation factor of the length, so that the equation can be used for different growth stages. Due to the segmentation results of the Mask R-CNN, an average precision of 94.1 % from the calculated ground truth value was achieved. The deviations could be caused by damaged leaves as well as by an incorrect estimation of the leaf length when the leaf end piece was not visible. Comparing the segmentation results of the leaves of 0.47 mAP with the result of the leaf area calculation of 94.1 %, it can be seen that due to the reduction of the leaf area calculation to the leaf length, smaller deviations in the segmentation do not have a strong effect on the leaf area calculation.

Table 2. Comparison of measured, calculated (calc.) and detected leaf area and the mean precision (MP) of the visible Leaves (vis.).

| Plant | Leaves total number | Measured LA (m$^2$) | Leaves visible number | Measured LA vis. (m$^2$) | Calc. gTruth (m$^2$) | Leaves detected number | Detected LA (m$^2$) |
|---|---|---|---|---|---|---|---|
| 1 | 15 | 0.66 | 13 | 0.57 | 0.61 | 12 | 0.62 |
| 2 | 17 | 0.81 | 15 | 0.71 | 0.80 | 13 | 0.74 |
| 3 | 17 | 0.82 | 14 | 0.67 | 0.70 | 14 | 0.72 |
| 4 | 18 | 0.84 | 16 | 0.71 | 0.92 | 16 | 0.99 |
| 5 | 16 | 0.94 | 11 | 0.60 | 0.66 | 11 | 0.66 |
| 6 | 17 | 0.96 | 15 | 0.85 | 0.70 | 13 | 0.62 |
| 7 | 16 | 0.76 | 13 | 0.62 | 0.63 | 13 | 0.64 |
| 8 | 19 | 1.04 | 15 | 0.82 | 0.74 | 13 | 0.65 |
| 9 | 15 | 0.76 | 12 | 0.61 | 0.59 | 12 | 0.61 |
| 10 | 15 | 0.67 | 15 | 0.67 | 0.61 | 11 | 0.54 |
| MP (%) | | | | | 89.8 | | 94.1 |

With a precision of the detected leaf area from the actual leaf area, including all leaves, of 86% and a precision of the detected cabbage volume to the real one of 94.6 %, further adjustments to the calculation process of the LA including the non-visible leaves are necessary in future research. Further, the method described would have to be tested using different growth stages and cabbage varieties to investigate the adaptability of the method.

**Conclusion**

A neural network-based instance segmentation as well as a SfM method for depth calculation were used to calculate the volume and total leaf area of individual cabbage plants with only one camera. Using a developed equation, the total leaf area could be calculated with 89.8% accuracy and the cabbage volume with 92.5% accuracy. By using the leaf length to calculate the leaf area, overlapping leaf areas could also be included in the total leaf area. The results presented in this paper refer to only one growth stage. Further analyses on the adaptability of the method to different growth stages and the use of different depth sensors are still needed.

**Acknowledgements**

The DiWenkLa-project supported this research by funds of the Federal Ministry of Food and Agriculture (BMEL) based on a decision of the Parliament of the Federal Republic of Germany via the Federal Office for Agriculture and Food (BLE) under the innovation support programme.